\documentclass[journal]{IEEEtran}
\usepackage{amsmath,amsfonts}
\usepackage{array}
\usepackage[caption=false,font=normalsize,labelfont=sf,textfont=sf]{subfig}
\usepackage{textcomp}
\usepackage{stfloats}
\usepackage[authoryear]{natbib}

\usepackage{url}
\usepackage{verbatim}
\usepackage{graphicx}
\usepackage{tabularx}
\usepackage{amsmath}
\usepackage{amsmath}
\usepackage{mathtools}
\usepackage{cite}
\usepackage{hyperref}

\usepackage{algorithm}
\usepackage{algpseudocode}

\begin{document}

\title{Artificial Disfluency Detection, Uh No, Disfluency Generation for the Masses}

\author{Tatiana Passali, Thanassis Mavropoulos, Grigorios Tsoumakas, \IEEEmembership{Senior Member, IEEE}, \\
Georgios Meditskos and Stefanos Vrochidis% <-this % stops a space
\IEEEcompsocitemizethanks{\IEEEcompsocthanksitem T. Passali is with the School of Informatics, Aristotle University of Thessaloniki, Thessaloniki, Greece and also with the Centre for Research and Tecnhology Hellas, Thessaloniki, Greece (e-mail: scpassali@csd.auth.gr and tpassali@iti.gr)\protect \\
% E-mail: see http://www.michaelshell.org/contact.html
\IEEEcompsocthanksitem T. Mavropoulos is with the Centre for Research and Tecnhology Hellas, Thessaloniki, Greece (e-mail: mavrathan.iti.gr)\protect \\
\IEEEcompsocthanksitem G. Tsoumakas is with the School of Informatics, Aristotle University of Thessaloniki, Thessaloniki, Greece (e-mail: greg@csd.auth.gr)\protect \\
\IEEEcompsocthanksitem G. Meditskos is with the the School of Informatics, Aristotle University of Thessaloniki, Thessaloniki, Greece and also with the Centre for Research and Tecnhology Hellas, Thessaloniki, Greece (e-mail: gmeditsk@csd.auth.gr and gmeditsk.iti.gr)\protect \\
\IEEEcompsocthanksitem S. Vrochidis is with the Centre for Research and Tecnhology Hellas, Thessaloniki, Greece (e-mail: stefanos.iti.gr).}}



\maketitle

\begin{abstract}
Existing approaches for disfluency detection typically require the existence of large annotated datasets. However, current datasets for this task are limited, suffer from class imbalance, and lack some types of disfluencies that can be encountered in real-world scenarios. This work proposes LARD, a method for automatically generating artificial disfluencies from fluent text. LARD can simulate all the different types of disfluencies (repetitions, replacements and restarts) based on the reparandum/interregnum annotation scheme. In addition, it incorporates contextual embeddings into the disfluency generation to produce realistic context-aware artificial disfluencies. Since the proposed method requires only fluent text, it can be used directly for training, bypassing the requirement of annotated disfluent data. Our empirical evaluation demonstrates that LARD can indeed be effectively used when no or only a few data are available. Furthermore, our detailed analysis suggests that the proposed method generates realistic disfluencies and increases the accuracy of existing disfluency detectors.
\end{abstract}

\begin{IEEEkeywords}
Disfluency Detection, Data Augmentation
\end{IEEEkeywords}

\section{Introduction}
\label{sec:introduction}
\IEEEPARstart{A}{utomatic} Speech Recognition (ASR) technology has recently achieved remarkable progress and is now an integral part of a wide range of intelligent systems, such as general-domain virtual assistants and specialized spoken dialogue systems for healthcare~\citep{latif2020speech}, migration~\citep{wanner2021towards}, finance~\citep{wang2020two} and education~\citep{maxwelll2021developing}. An important challenge when deploying such technology in a real-world application is {\em speech disfluencies}, such as filled pauses, self-repairs, repetitions, hesitations, and false starts. These linguistic phenomena are a common feature of spontaneous human speech~\citep{shriberg1996disfluencies}.

The presence of speech disfluencies can have a negative impact not only on the readability of generated transcripts, but also on the performance in downstream tasks, such as machine translation, question answering, and summarization. For example, \citet{gupta2021disfl} have shown that the presence of disfluencies in questions can significantly drop the performance of a question-answering model. Existing conversational systems have a varying degree of robustness against speech disfluencies, either handling such phenomena implicitly as part of a downstream task model~\citep{jamshid2020end}, or explicitly as a post-processing step after the ASR component and before a downstream task model~\citep{Zayats2016Disfluency, jamshid2018disfluency}. Recent literature has shown that the latter approach can significantly improve the performance of ASR systems, bringing them closer to human-level performance~\citep{jamshid2020end}.

Several methods for explicitly handling disfluencies with a separate component have been proposed in the literature. The most common is sequence tagging~\citep{ostendorf2013sequential, Liu2006enriching, ferguson2015disfluency}, where each token of the input sequence is classified as fluent or disfluent. The advent of deep learning (DL) led to even more powerful models for disfluency detection, varying from recurrent neural networks (RNNs)~\citep{Hough2015incremental, Zayats2016Disfluency} and convolutional neural networks (CNNs)~\citep{jamshid2018disfluency} to more recent Transformer-based methods~\citep{dong2019adapting, Wang_Che_Liu_Qin_Liu_Wang_2020, rocholl2021disfluency}.

Despite the great progress of these models, they still suffer from a significant limitation: they heavily rely on the existence of annotated datasets for training and evaluation. However, existing datasets for disfluency detection are limited and do not cover sufficiently all the different types of disfluency that can be encountered in real-world scenarios. For example, Switchboard~\citep{Godfrey1992Switch}, the largest and most commonly used dataset for disfluency detection, has a highly imbalanced distribution among the different disfluency classes, with more than 50\% of the examples belonging to repetitions~\citep{shriberg1996disfluencies}, which is the easiest class of disfluencies according to the literature~\citep{Zayats2016Disfluency, Zayats2019Switch}. Training models on Switchboard can lead to a significant bias towards the repetition class, while evaluating models on it will lead to misleading results with respect to their real-world performance~\citep{passali2022lard}. An option to remedy this drawback would be to manually annotate datasets with more complex disfluencies like replacements and restarts. However, this would require enormous human labeling effort and appropriately trained personnel.

Motivated by this observation, we propose an approach for automatically generating highly realistic context-aware artificial disfluencies from existing fluent dialogue corpora, without any human supervision. %Our method can effectively tackle the aforementioned limitation since it can generate large-scale realistic context-aware disfluencies without any human supervision. 
These artificial disfluencies can be used directly for training any disfluency detection model, contrary to existing augmentation techniques which are used only for self-supervised pre-training~\citep{dong2019adapting, yang2020planning, Wang_Che_Liu_Qin_Liu_Wang_2020} via a denoising objective. In particular, we propose three distinct algorithms, one for each of the three most common categories of disfluencies: repetitions, replacements, and restarts. To the best of our knowledge, this is the first work that takes into account context-based representations for artificially labeling data in disfluency detection. 

A preliminary version of this work was presented in~\citet{passali2022lard}. Here, we extend our previous approach by: a) incorporating context-aware representations for generating more complicated disfluencies such as replacements, b) refining the previously proposed algorithms to improve the quality of the generated disfluencies, and c) providing an in-depth experimental evaluation using an additional dataset, both in a regular and in a low-resource setup. In particular, the replacement algorithm works by first extracting possible repair candidates for the replacement using semantic relations such as hypernyms and hyponyms and then selecting the best repair candidate using a BERT-based language model. This way, we ensure that the inserted replacement does not affect the semantic relation between the fluent and the generated disfluent sequence. In addition, we appropriately modify the repetition and restart algorithms to filter non-realistic results.

Our contributions can be summarized as follows:
\begin{itemize}
    \item We propose a method for automatically generating realistic artificial repetitions, replacements, and restarts.
    \item We provide an empirical evaluation of the proposed method on Switchboard and Disfl-QA. Results show that our method can boost the performance of existing models for disfluency detection.
    \item We conducted a proof-of-concept experiment in a low-resource setup, demonstrating that our method can be successfully used with few or no data at all. This is especially important for low-resource languages where disfluent data is scarce.
\end{itemize}

The rest of this paper is organized as follows.  Section~\ref{sec:background} discusses the structure of disfluencies in human dialogue. Section~\ref{sec:related_work} reviews related literature. Section~\ref{sec:proposed_method}, introduces the proposed method. Section~\ref{sec:experiments} presents and discusses the results of our empirical evaluation. Finally, Section~\ref{sec:conclusion}, concludes our work and points to interesting future research directions.

\section{Disfluencies in Human Dialogue}
\label{sec:background}
This section introduces the structure, as well as the different categories of disfluencies.

\subsection{Structure of Disfluencies}
\citet{shriberg1994preliminaries} introduced a standard annotation scheme, called \textit{reparandum/interregnum}, for identifying disfluencies. This annotation scheme involves the following fragments: a) the \textit{reparandum}, b) the \textit{interruption point}, c) the \textit{repair}, and, optionally, d) the \textit{interregnum}, which is located right before the repair. 

The reparandum indicates the disfluent part of the utterance: the part that is not correct and must be replaced or ignored. Typically, the speaker attempts to rephrase, edit or restate the reparandum. The reparandum is usually short, involving 2 to 3 words. Even though the reparandum is usually considered as a ``rough'' copy of the repair, it can also be completely irrelevant to the repair.

The interruption point indicates the start of the repair, if no interregnum exists. If an interregnum exists, then the interruption point indicates the start of the interregnum, after which the repair follows. The interruption point does not indicate an actual word, it rather marks the moment of speech when a speaker realizes the error and initiates the correction process. In other words, the interruption point typically occurs right after the last word that is being told before the interruption of speech. 

The interregnum consists of repair cues, such as those shown in Table~\ref{tab:repair_cues_types}, %filled pauses (e.g., ``um'', ``uh'') or fixed words (e.g., ``you know'', ``i mean'') 
that are typically used to fill the gap between the disfluent speech and the associated repair. These types of fillers are generally ignored from dialogue systems as they do not contain any useful information. Most of the time, the presence of a repair cue indicates the presence of a disfluency in the utterance. %Some examples of repair cues are shown in Table~\ref{tab:repair_cues_types}. 
Interregnums are relatively easy to detect as they are usually fixed phrases, in contrast to reparandums that require a deeper understanding of the complex dialogue flow. For this reason, many methods ignore the interregnum and focus only on the detection of the reparandum and the repair~\citep{Zayats2016Disfluency,Zayats2019Switch,jamshid2018disfluency,Bach2019NoisyBM}.

\begin{table}[hbt]
\caption{Different types of repair cues as described by\protect\citet{shriberg1994preliminaries}}
\label{tab:repair_cues_types}
\centering
\begin{tabularx}{\columnwidth}{|l|X|}
\hline
\textbf{Repair cues} & \textbf{Examples} \\ \hline
filled pauses & um, uh   \\ \hline
editing phrases    & oops, no, sorry, wait, I meant to say \\  \hline
discourse markers &  well, actually, okay, you know, I mean \\
\hline
\end{tabularx}
\end{table}

We typically meet a repair in a speaker’s utterance either after the presence of an interregnum or directly after the interruption point. The repair is the corrected fragment of speech and is usually different from the reparandum. However, in some cases the repair can be exactly the same as the reparandum or even completely empty.

\begin{figure}
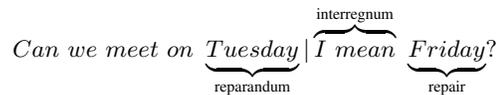

    \centering
    \small
    \[    Can \ we \ meet \ on  \ \underbrace{ Tuesday}_{%
   \mathclap{\text{reparandum}}} \vert  \! \! \  \overbrace{I \ mean}^{\substack{\text{interregnum}}} \ 
   \underbrace{ Friday}_{%
   \mathclap{\text{repair}}} ? \] 
    \caption{An example of a disfluency, where the reparandum, interregnum and repair are illustrated. The vertical bar indicates the interruption point.}
    \label{fig:disfluency_example}
\end{figure}

An example of the aforementioned annotation scheme is described in Fig.~\ref{fig:disfluency_example}, where the speaker mistakenly says Tuesday instead of Friday. The disfluent part of this example consists of the words \textit{``on Tuesday''} (reparandum), which is followed by an interregnum ({\em ``I mean''}), and eventually by the correct choice of day (\textit{``on Friday''}) that makes the particular repair. To simplify the presentation of disfluencies under this scheme, the following notation is typically used to indicate the different parts of disfluencies:
\[ \text{[reparandum + \{interregnum\} repair] }\]
In this notation, the brackets (``['' and ``]'') are used to indicate the disfluency along with the repair.  The interruption point is indicated by the plus symbol (``+''), while the interregnum, if present, is indicated by curly brackets (``\{'', ``\}'').

\subsection{Categories of Disfluencies}
\label{sec:categories}
According to the speech repair typology of~\citet{shriberg1994preliminaries}, disfluencies can be categorized into three distinct classes, as described below:
\begin{itemize}
\item \textbf{Repetitions}: The speaker repeats a word, a phrase or a sequence of words. In repetitions, the reparandum and the corresponding repair are actually the same. This type of disfluency is the most common and is particularly easy to be detected~\citep{Zayats2014MultidomainDA,Zayats2019Switch}. 
\item \textbf{Replacements}: The speaker replaces the disfluent word(s) or phrase with the fluent one. In replacements, the reparandum is replaced by the repair.
\item \textbf{Restarts}: The speaker abandons completely the initial utterance and restarts it. This type of disfluency does not actually involve a repair as the speaker begins a completely new utterance. 
\end{itemize}

Table~\ref{tab:disfluency_types} presents an example for each of these three categories of discfluencies.

\begin{table}[hbt]
\caption{Different types of disfluencies annotated based on the reparandum/interregnum scheme.}
%\textit{Repetitions}: The reparandum is same with the associated repair. \textit{Replacements}: The reparandum is replaced with the associated repair after the interruption point. \textit{Restarts}: The repair fragment is empty, since the initial utterance is completely abandoned.}
\label{tab:disfluency_types}
\centering
\resizebox{\columnwidth}{!}{%
\begin{tabular}{|l|l|}
\hline
\textbf{Type}       & \textbf{Example}  \\ \hline
repetition & Let's meet [today + today].     \\
replacement    & I want [the blue + \{no\} the red ] one.     \\
restart    & [Why don't you + ] I will do it later. \\
\hline
\end{tabular}%
}
\end{table}

\section{Related Work}
\label{sec:related_work}
This section reviews related work on disfluency detection. First, we provide an overview of existing models and methods for detecting disfluencies. Then, we present datasets that are used to train the aforementioned models. Finally, we discuss existing augmentation techniques that have been proposed to address the issue of limited data for disfluency detection.

\subsection{Disfluency Detection Models}
Several approaches have been proposed for handling speech disfluencies in a dialogue. These approaches are divided into four significant categories: a) \textit{sequence tagging}, b) \textit{translation-based}, c) \textit{parsing-based} and d) \textit{noisy-channel} methods. Sequence tagging, which is the most common approach for disfluency detection, classifies each token of the input sequence as fluent or disfluent. This approach includes a wide range of different models varying from  Hidden Markov Models~\citep{Liu2006enriching}, Conditional Random Fields (CRFs)~\citep{Georgila2009,ostendorf2013sequential}, Semi-Markov CRFs~\citep{ferguson2015disfluency}, CNNs~\citep{jamshid2018disfluency}, RNNs~\citep{Zayats2016Disfluency}, to more recent Transformer-based architectures ~\citep{dong2019adapting,Wang_Che_Liu_Qin_Liu_Wang_2020, chen2022teaching}, and even using an on-device lightweight setup~\citep{rocholl2021disfluency}. On the other hand, translation-based methods approach disfluency detection as a sequence-to-sequence problem using encoder-decoder models to generate the corrected sequence without the disfluent fragments by ``translating'' the disfluent input into a fluent sequence~\citep{saini2021disfluency}. Parsing-based approaches detect concurrently both the syntactic structure and the disfluencies of the input sequence~\citep{wang2017transition, tran2019parsing}. These approaches require corresponding annotated training datasets that include both syntactic and disfluency annotations. Noisy-channel models~\citep{charniak2001edit,johnson2004noisy} detect disfluencies by computing the similarity between the reparandum and the repair.

\subsection{Disfluency Detection Datasets}
\noindent \textbf{Switchboard}~\citep{Godfrey1992Switch} is the most popular disfluency detection dataset. It is the largest human-annotated dataset with real disfluencies at token-level, based on the reparandum/interregnum scheme. Even though Switchboard contains approximately 190,000 multi-speaker transcriptions of English dialogues on different topics, it has a highly imbalanced class distribution, with only 5.9\% of tokens being disfluent. In addition, more than 50\% of these disfluent tokens belong to simple repetitions~\citep{charniak2001edit}, which has been shown to be the most trivial to tackle disfluency type~\citep{charniak2001edit}.
\newline

\noindent \textbf{Disfl-QA}~\citep{gupta2021disfl} is a recent dataset for disfluency detection containing 12,000 human-annotated disfluent questions in English. In contrast to Switchboard, Disfl-QA has a higher number of more complicated disfluencies, such as replacements and restarts. However, it is significantly smaller. Furthermore, because it lacks token-level annotations, it can only be used directly by translation-based models.
\newline

\noindent \textbf{Fisher English Training Transcripts}~\citep{cieri2004fisher} contains approximately 1.3 million transcriptions of real dialogues between speakers. Despite being larger than Switchboard, it lacks annotations for the disfluencies it contains and thus cannot be readily used for the task of disfluency detection.

\subsection{Data Augmentation Techniques}
Despite the significant progress of DL models, they still rely heavily on the quality and quantity of training data. To address the issue of limited datasets, some early steps towards data augmentation have been made, varying from semi-supervised techniques~\citep{wang2018semi} to self-training approaches~\citep{jamshid2020improving}, either by annotating unlabeled datasets, such as Fisher, using existing disfluency detectors trained on Switchboard, or by unsupervised pre-training techniques using artificial data~\citep{dong2019adapting, yang2020planning, Wang_Che_Liu_Qin_Liu_Wang_2020}. Our work is closer to the latter approach.

\citet{dong2019adapting} corrupt the input sequence for pre-training by adding noise based on randomly selected input words, while a similar approach is adopted by~\citet{yang2020planning}, where a simple planner-generator model is used to generate and place disfluent text in a fluent sentence. \citet{Wang_Che_Liu_Qin_Liu_Wang_2020} generate artificial disfluencies by randomly repeating, inserting, and removing words. As the inserted tokens are randomly selected, the semantic consistency between the created disfluent sentence and the fluent one cannot be guaranteed. The aforementioned approaches rely on simple rules and techniques that are incapable of generating all the different kinds of complex disfluencies that occur in real-world scenarios. Our work differs from these approaches in several aspects. First, it is carefully designed to generate and handle all types of disfluencies. In addition, the proposed method incorporates contextual representations for the generation of synthetic disfluencies, leading to disfluencies that are closer to the ones occurring in natural dialogue flows. Finally, the proposed method can be used to generate annotated disfluencies for directly training supervised disfluency detectors, instead of defining a denoising pre-training objective, which can be computationally costly.

\section{Proposed Method}
\label{sec:proposed_method}
In this section, we introduce our improved Large-scale ARtificial Disfluency (LARD) generation method for synthesizing artificial disfluencies from fluent text. The proposed method consists of three algorithms for supporting the corresponding types of disfluencies, namely repetitions, replacements, and restarts.

Given any set of fluent sentences, we can use these algorithms to generate a new set of annotated disfluent sentences containing artificial repetitions, replacements, and restarts at any ratio. Even though all the algorithms can consume the same fluent sentence to generate disfluencies, we use different sub-sets of the fluent set for each algorithm to increase the variation of the generated examples.

\subsection{Repetition Algorithm}
%The repetition algorithm can be used to generate realistic artificial repetitions given any fluent sentence. 
The procedure used for generating artificial repetitions is summarized in Alg.~\ref{alg:repetitions}. This algorithm works by randomly repeating words or sequences of words based on a given degree ranging from 1 to 3.

For example, given the fluent sequence ``{\em Thank you for your help}'', we can simulate an artificial disfluent sequence that includes a degree 1 repetition by repeating the word ``for'' in the third position, or a degree 2 repetition by repeating the sequence of words ``thank you'', starting from the first position as follows: 

\begin{center}
Degree 1: Thank you [for + for] your help
\end{center}
\begin{center}
Degree 2: [Thank you + thank you] for your help
\end{center}

% \begin{equation*}
% \begin{array}{l}
%   S_{d} \text{ (degree 1)} = \text{``Thank you [for + for] your help'',} \\ \\
%    S_{d} \text{ (degree 2)} = \text{``[Thank you + thank you] for your help''.}
%  \end{array}
%\end{equation*}

During this procedure, we ensure that we select only valid candidates for repetition by excluding tokens that belong to punctuation marks. The repetition algorithm can be used for generating up to degree 3 repetitions, but it can also be easily modified to generate repetitions of higher degrees according to the length of the input sequence.

\begin{algorithm}[t]
\caption{Repetition algorithm}\label{alg:repetitions}
\begin{algorithmic}[1]
\Require{A fluent sequence $S_f$, the degree of repetition $d_r \in\{1,2,3\}$}
\Ensure{A disfluent sequence $S_{d}$ which contains a repetition based on given degree $d_r$ }
\Procedure{Generate Repetitions}{$S_f,d_r$}
    \State $l_{s} \gets length(S_{f})$
    \State Choose $random_{idx} \gets uniform(0, l_{s}-d_r$) matching the condition that the word of the selected index, as well as words up to degree $d_r$ of the subsequent indexes, do not belong to punctuation marks.
    \State $S_{d} \gets$ Repeat the subsequence of tokens in $S_{f}$ starting from index $random_{idx}$ to degree $d_r$.
% % \For{$i \gets 0$ to $d_r$} $S_{dis} \gets$ Repeat the token in the index $random_{idx} + i $ of $S_f$
% \EndFor
\State \Return $S_{d}$
\EndProcedure
\end{algorithmic}
\end{algorithm}

\subsection{Replacement Algorithm}
%The replacement algorithm can be used for generating artificial replacements. 
To preserve the semantic consistency between the initial fluent sentence and the generated disfluent one, we introduce a two-step replacement candidate selection process exploiting the semantic relation of the selected candidate with hypernyms and hyponyms. Then, a BERT-based language model is used to select the best candidate from a set of possible candidates. The replacement algorithm is described in Alg.~\ref{alg:replacements}. The steps for generating artificial replacements can be summarized as follows:

\begin{enumerate}
    \item First, we extract a random word of the initial fluent sentence, whose part-of-speech (POS) is either noun, verb, or adjective, to serve as our \textit{repair candidate}. The POS tagger of NLTK~\citep{bird2009natural} is used in this step.
    \item We detect the hypernym class of the repair candidate, namely its broader semantic category, using NLTK~\citep{bird2009natural}. In case more than one hypernym exist, we extract all the possible hypernyms.
    \item For each hypernym, apart from the original candidate, we extract $N$ hyponyms, namely words that belong to the same broader semantic category. Since hyponyms share common semantic relations, we assume they will be semantically close to the initial repair candidate. 
    \item For each hyponym, we generate a new sentence based on the initial sentence by substituting the selected repair candidate with the corresponding hyponym. Then, we compute the similarity between all the generated sentences and the original sentence.
    \item Instead of randomly selecting a hyponym from the extracted set of hyponyms, we select the best candidate, the one in the sentence with the highest similarity. The best candidate serves as our \textit{reparandum candidate}.
    \item To simulate an artificial replacement, the reparandum candidate is placed right before the repair candidate in the initial sentence, to ensure that the fluent sentence will be realistic. The repair candidate is always one word, but the reparandum candidate can vary from 1 to 4 words according to the extracted hyponym.
    \item Optionally, we can add a repair cue between the replacement and the repair candidate. We create a fixed list of repair cues with some variations of words and phrases from Table~\ref{tab:repair_cues_types}. Note that filled pauses, such as um and uh, are not included in this list as these types of disfluencies are typically removed automatically from recent ASR systems.
\end{enumerate}

\begin{algorithm}[t]
\caption{Replacement algorithm}\label{alg:replacements}
\begin{algorithmic}[1]
\Require{A fluent sequence $S_f$, part-of-speech $s_{pos} \in$ \{noun, verb, adjective\}, the range of hyponyms $N$, a boolean variable $cue$,}
\Ensure{A disfluent sequence $S_{d}$ which contains a replacement based on the part-of-speech $s_{pos}$}
\Procedure{Create Replacement}{$S_f$, $s_{pos}$, $N$, $cue$}
\While{there is $s_{pos}$ in $S_f$}
\State $W_{pos} \gets $ a list with all of the available words of the sequence $S_f$ that belong to the selected input part-of-speech $s_{pos}$
\State Select randomly one of the available words from $W_{pos}$ to serve as repair candidate 
\State Detect the hypernym of the repair candidate. If there are more than one hypernym extract all the possible hypernyms.
\State Extract N hyponyms for each hypernym, excluding the repair candidate.
\State $\{S\} \gets N$ new sequences based on $S_f$ with the substitution of the repair candidate with the corresponding hyponym in $S_f$
\State Compute the similarity between all the sentences and the $S_f$ and select as reparandum candidate the hyponym whose sentence has the highest similarity. 
\State $d_{r} \gets uniform(0, 3)$ 
\State Repeat $d_{r}$ tokens before the replacement candidate
\State Place the reparandum candidate
\If{$cue$ == True} randomly select a repair cue from list and place it after the reparandum candidate.
\EndIf
\State Continue the rest of $S_f$.
\State \Return{$S_d$}
\EndWhile
\EndProcedure
\end{algorithmic}
\end{algorithm}

To compute the similarity between each generated sentence with the initial sentence, we first employ a BERT-based model to extract sentence embeddings $SE$, by mean-pooling all the token embeddings of the given sentence as follows:

\begin{equation}
SE = \frac{1}{n}\sum_{i=0}^{n-1} TE(i),
\end{equation}
where $TE(i)$ represents the token embedding of the $i^{th}$ token, while $n$ denotes the number of tokens in the given sentence.

The final reparandum candidate $r$ is selected as follows:

\begin{equation}
r = \max\limits_{i \in T} \{ sim(SE_{s}, SE_{i}) \},
\end{equation}
where $T$ represents the set of the hyponym-generated sentences, $SE_{s}$ and $SE_{i}$ denote the sentence embeddings of the initial sentence $s$ and the generated sentence with the hyponym $i$ of the set $T$ respectively, while $sim$ denotes the cosine similarity between two vectors $\mathbf{x}_y$ and $\mathbf{x}_z$, computed as follows:

\begin{equation}
sim(\mathbf{x}_{y}, \mathbf{x}_{z}) = \frac{\mathbf{x}_{y}, \mathbf{x}_{z}}{\|\mathbf{x}_{y}\| \|\mathbf{x}_{z}\|}.
\end{equation} 

The replacement algorithm can generate simple replacements involving only one word, but can also be used to simulate more complicated $n$-word replacements ($n > 1$) involving multiple words in both reparandum and repair. For the latter, we repeat a sequence of words before the reparandum candidate. We limit this sequence length to 3 words in order to ensure that the generated sentences will be coherent and realistic. 

For example, given fluent sequence ``{\em I would like to eat pancakes for breakfast}'', we can simulate an artificial noun-replacement by selecting the word ``pancakes'', whose POS is a noun, as our repair candidate. We extract its hypernym class, namely ``cake''. Some possible hyponyms for the word ``cake''  are ``gingerbread'', ``cheesecake'', ``doughnut'', ``honey cake'', ``brownie'' etc. Then, we generate all the possible sentence candidates, like ``{\em I would like to eat gingerbread for breakfast}'' and ``{\em I would like to eat cheesecake for breakfast}'', and compute their similarity with the initial fluent sentence. The best candidate e.g., ``cheesecake'', is the one belonging to the sentence with the highest similarity. Note that we can also use a repair cue from the fixed list of repair cues e.g., ``no wait". Finally, we can generate a simple artificial disfluent sequence by replacing the reparandum candidate ``cheesecake'' with the repair candidate ``pancakes'' or a three-word replacement by also repeating the sub-sequence of words ``to eat'', starting from the fourth position as follows:  

\begin{center}
\qquad One-word: I would like to eat [cheesecake + \{no\} pancakes] for breakfast
\end{center}
\begin{center}
\qquad Three-word: I would like [ to eat cheesecake + \{no\} to eat pancakes] for breakfast 
\end{center}

% \begin{equation*}
% \begin{array}{l}
%   S_{d} \text{ (1-word)} = \text{``I would like to eat [ cheesecake + \{no\}} \\ \text{pancakes] for breakfast .'',} \\ \\
%    S_{d} \text{ (3-word)} = \text{``I would like [ to eat cheesecake + \{no\}} \\ \text{to eat pancakes] for breakfast .''.}
%  \end{array}
%\end{equation*}

Following this procedure, we can simulate six different sub-classes of replacements: noun replacement with and without repair cue, verb replacement with and without repair cue and adjective replacement with and without repair cue.

\subsection{Restart Algorithm}
Contrary to the repetition and replacement algorithm, this algorithm assumes the existence of a set of two or more fluent sentences. The steps for generating artificial restarts are summarized in Alg.~\ref{alg:restarts}. First, we extract two different sentences from a given fluent set. Then, we split the first sentence at a random position and we concatenate this broken sentence with the second fluent one. 

\begin{algorithm}[t]
\caption{Restart algorithm}\label{alg:restarts}
\begin{algorithmic}[1]
\Require{Two fluent sequences $S_{f1}$ and $S_{f2}$, with $S_{f1} \neq S_{f2}$}
\Ensure{A disfluent sequence $S_{d}$}
\Procedure{Create Restarts}{$S_{f1}$, $S_{f2}$}
\State $S'_{f1} \gets$ $S_{f1}$ broken in a random position. 
\While{$S'_{f1} \neq$ beginning of $S_{f2}$}
\State Check for connective words at the end of $S'_{f1}$ or at the beginning of $S_{f2}$
\If{no connective words detected}
\State $S_{d} \gets join(S'_{f1}, S_{f2})$
\State \Return $S_{d}$
\Else \State Select another sequence $S_{f2}$
\EndIf
\EndWhile
\EndProcedure
\end{algorithmic}
\end{algorithm}

For example, given two fluent sequences ``{\em I would like to buy a new dress}'' and ``{\em Can we meet on Tuesday?}'', we can simulate a disfluent sentence with an artificial restart by splitting the first sentencein the second position and concatenating it with the second sentence as follows:
\begin{center}
    [I would +] can we meet on Tuesday?
\end{center}
%\begin{equation*}
%    S_{d} = \text{``[I would +] can we meet on Tuesday?''} \\
%\end{equation*}

As discussed in Section~\ref{sec:background}, this type of disfluency does not involve a repair as the speaker abandons the previous sentence and restarts with a new one. 

During this procedure, it is possible to accidentally generate a fluent sentence instead of a disfluent one. This might happen when the connection point for the given sentences happens to belong to a connective word or phrase. As connective words are typically used to link one sentence with another, their presence might result in a fluent and coherent sentence. To alleviate this issue, we create a list of possible connectives according to the different types that can occur in a real dialogue flow, e.g. additive (and, also etc.), causal (because, if, since etc.), adversative (but, however, etc.), and temporal (after, before, etc.).
% as shown in Table~\ref{tab:connective_words}. 
Then, while concatenating the two sentences, we check for the presence of connective words and filter such sentences to avoid the generation of fluent examples. Also, we ensure that the end of the first broken sentence is not the same as the beginning of the second sentence to avoid unintentionally generating a repetition instead of a restart.

% \begin{table}[t]
% \caption{Different types of connective words as indicated by~\protect\citet{halliday1976}}
% \label{tab:connective_words}
% \centering
% \begin{tabularx}{\columnwidth}{|l|X|}
% \hline
% \textbf{Connective Type} & \textbf{Examples} \\ \hline
% Additive & and, also, as well as, in addition, for instance, furthermore, for example   \\ \hline
% Causal   & because, if, even if, since, so, therefore, consequently, that`s why \\  \hline
% Adversative &  but, however, although, even though, or else, yet, instead, at least, anyhow \\ \hline
% Temporal &  after, before, next, soon, finally, then, now, when \\

% \hline
% \end{tabularx}
% \end{table}

\section{Empirical Evaluation}
\label{sec:experiments} 
In this section, we introduce the training and evaluation details of our experimental setup, as well as present and discuss the results of our empirical evaluation.

\subsection{Experimental Setup}

We conduct experiments with sequence tagging and translation-based approaches. For sequence tagging, we use the pre-trained BERT-base uncased model, which consists of 12 encoder layers with 12 attention heads. For the translation-based approach, we use the pre-trained T5-base model~\citep{RaffelExploring2020}, which consists of 12 encoder and decoder layers with 12 attention heads as well. In addition, for extracting vector representations when generating artificial replacements, we use the multi-qa-distilbert-cos-v1 model~\citep{reimers2019sentence}, which is a lightweight distillBERT-based model fine-tuned appropriately for the task of semantic search.

We fine-tune all the models for 10 epochs with learning rate 2e-5 and batch size 16. We do not perform extensive hyper-parameter search as this lies beyond the scope of this paper. Further fine-tuning or more extensive hyper-parameter tuning could lead to improved models. All the conducted experiments were performed using the Hugging Face library\footnote{\href{https://huggingface.co/}{https://huggingface.co/}}.

We train and evaluate our disfluency models on Switchboard and Disfl-QA. Switchboard provides token-level annotations and can be easily used for both sequence-tagging and translation-based approaches. For sequence tagging, we classify each token as fluent or disfluent. More specifically, we classify as disfluent all the tokens that are located before the repair inside the disfluent fragment. Similarly, we classify as fluent all the tokens after the interruption point inside the disfluent fragment as well as all the tokens located outside the disfluent fragment. We use the established train/validation/test split for training and evaluation~\citep{charniak2001edit}. Following~\citep{rocholl2021disfluency}, we also discard partial words and words that belong to interregnum such as ``uh'', ``um'', etc. because they are trivially detected by recent ASR systems. 

Disfl-QA provides only raw disfluent sentences accompanied with their respective target fluent versions. Therefore, it can only be used with translation-based models. We adapt this dataset for sequence tagging by mapping one-to-one annotations between the fluent and disfluent sentences. Examples where fluent sentences are not sub-sentences of the disfluent ones are discarded. Some statistics for both datasets are shown in Table~\ref{tab:dataset_statistics}.

%\noindent \textbf{Evaluation Metrics} For all the sequence tagging models, we report token-based recall, precision and F-measure. For translation-based models, we report BLEU score~\citep{papineni2002bleu}. 
%\newline

%\noindent \textbf{Models and Training Details} For all the conducted experiments, we use Transformer-based models as follows:

%\begin{itemize}
%    \item For the \textit{sequence tagging} task, we use the pre-trained BERT-base uncased model which consists of 12 encoder layers with 12 attention heads. 
%    \item For the \textit{translation-based} task, we use the pre-trained T5-base model~\citep{RaffelExploring2020} which consists of 12 encoder and decoder layers with 12 attention heads as well. 
%    \item For extracting the vector representations when generating artificial replacements, we use the multi-qa-distilbert-cos-v1 model~\citep{reimers2019sentence}, which is a lightweight distillBERT-based model fine-tuned appropriately for the task of semantic search.
%\end{itemize}

%We fine-tune all the models for 10 epochs with learning rate 2e-5 and 16 batch size. We do not perform extensive hyper-parameter searching as this lies beyond the scope of this paper. However, note that further fine-tuning or more extensive hyper-parameter tuning could potentially lead to improved models. All the conducted experiments were performed using the Hugging Face library~\citep{wolf2020transformers}.

\begin{table}[!hbt]
\caption{Dataset Statistics. All numbers are reported in sentences.}
\label{tab:dataset_statistics}
\resizebox{0.99\columnwidth}{!}{%
\begin{tabular}{cccc} \hline
\textbf{Dataset} & \textbf{Train} & \textbf{Validation} & \textbf{Test}     \\
\hline
Switchboard & 165,463 & 14,026 & 7,406            \\ 
Disfl-QA (original) & 7,182 & 1,000  & 3,643 \\ 
Disfl-QA (sequence tagging) & 5,537 & 780  & 2,460 \\ \hline
\end{tabular}}
\end{table}

For all the sequence tagging models, we report token-based recall, precision and F-measure. For translation-based models, we report BLEU score~\citep{papineni2002bleu}. 

\subsection{Results}
To demonstrate the effectiveness of the proposed method, we conduct two different types of experiments, examining: a) the effect of the proposed method in a low resource setup when no or few data are available, and b) the compatibility of the generated artificial disfluencies with real data.

% switchboard
% train: F: 1041859, D: 50186, fluent (sent.): 137191, disfl (sent.): 28272
% test: F: 51209, D: 3062, fluent (sent.): 5801, disfl (sent.): 1605
% validation F: 97847, D: 5527, fluent (sent.): 10998, disfl (sent.): 3028

% disfl-qa
% train: F: 52523, D: 25065, all sent. disfluent
% test: F: 24389, D: 11041, all sent. disfluent
% validation F:  7286, D: 3593, all sent. disfluent

\subsubsection{Low-resource setting} 
% As described in Section~\ref{sec:proposed_method}, LARD can be directly used to generate artificial disfluencies from fluent text without requiring any annotations or human effort. 
In this experiment, we simulate a low-resource setting by keeping only a few samples from each dataset according to their initial size. We use a sequence-tagging model for all these experiments. Our evaluation includes the following setups: 

% \begin{table*}[!t]
% \caption{Experimental results on models fine-tuned on samples of the Switchboard dataset, when no or few data are available. Average and per class token-based precision (prec), recall (rec) and f-measure (f1) (\%) is reported.}
% \tiny
% \label{tab:low_resource}
% \resizebox{0.99\textwidth}{!}{%
% \begin{tabular}{cccc|ccc|ccc} \hline 
% & \multicolumn{3}{c}{\textbf{Overall Scores}} 
% & \multicolumn{3}{c}{\textbf{Disfluent Class}} 
% & \multicolumn{3}{c}{\textbf{Fluent Class}} \\
% & \textbf{Prec} & \textbf{Rec} & \textbf{F1} & \textbf{Prec} & \textbf{Rec} & \textbf{F1} & \textbf{Prec} & \textbf{Rec} & \textbf{F1}  \\ \hline
% 1K real only & 92.60 & 76.46 & 82.41 & 87.94 & 53.36 & 66.42 & 97.27 & 99.56 & 98.40 \\ 
% 2K real only & 91.56 & 84.42 & 87.61 & 84.93 & 69.59 & 76.50 & 98.20 & 99.26 & 98.72 \\ 
% 5K real only & 91.23 & 84.63 & 87.60 & 84.24 & \textbf{70.05} & 76.49 & 98.22 & 99.22 & 98.71 \\ \hline
% artificial only  & 95.93 & 78.39 & 84.84 & 94.37 & 56.98 & 71.06 & 97.48 & 99.79 & 98.62  \\ 
% artificial + 1K real  & 96.98 & 80.40 & 86.76 & 96.24 & 61.10 & 74.75 & 97.72 & 99.85 & 98.77 \\ 
% artificial + 2K real       & 96.31 & 82.23 & 87.85 & 94.69 & 64.69 & 76.87 & 97.92 & 99.78 & 97.80 \\
% artificial + 5K real       & \textbf{98.08} & \textbf{84.81} & \textbf{89.71} & \textbf{94.81} & 69.85 & \textbf{80.44} & \textbf{98.22} & \textbf{99.77} & \textbf{98.99} \\  
% \hline
% \end{tabular}}
% \end{table*}

\begin{itemize}
\item \textbf{Only artificial samples (Switchboard).} We use LARD to generate different types of artificial disfluencies on the fluent set of Switchboard (82,315 fluent sentences). To get closer to the original distribution of the Switchboard test set ($\sim22\%$ disfluent sentences), we generate 31,034 examples of artificial disfluencies, consisting of 17,047 repetitions and 13,987 replacements. We do not include any artificial restarts because, as demonstrated later in this section, the nature of Switchboard dialogues can negatively affect the model's performance when the restart algorithm is used. We fine-tune the model on 82,315 fluent examples of Switchboard combined with the 31,034 generated artificial disfluencies, leading to an artificial training set of 113,349 examples. We do not include any real disfluency from the original dataset.

\item \textbf{Only artificial samples (Disfl-QA).}
We use the 5,537 target fluent sentences of Disfl-QA to generate different types of artificial disfluencies. Similar to Switchboard, we maintain the original distribution of disfluencies in the Disfl-QA dataset (more than 65\% replacements and 30\% restarts). After applying LARD, we obtain 3,600 artificial replacements and 1,937 artificial restarts. We do not include artificial repetitions since repetitions are not included in the original training set of Disfl-QA. We do not include any real disfluency from the original datasets. 

\item \textbf{1K/2K/5K real samples (Switchboard).} We sample 1,000 (841 fluent and 169 disfluent), 2,000 (1,682 fluent and 318 disfluent) and 5,000 (4,217 fluent and 783 disfluent) examples from the original Switchboard training dataset and we fine-tune the model on the corresponding subset. We keep the original distribution of the training data regarding the number of repetitions, replacements, and restarts.

\item \textbf{Artificial + 1K/2K/5K real samples (Switchboard).} We combine the samples of real examples of Switchboard with the corresponding generated artificial disfluencies.

\item \textbf{100/200/500 real samples (Disfl-QA).} We sample 100, 200 and 500 disfluent examples from the original Disfl-QA training and we fine-tune the model on the corresponding subset. Since Disfl-QA is a significantly smaller dataset than Switchboard, we reduce the samples of real examples by an order of magnitude. Note that Disfl-QA contains only the original disfluent examples. 

\item \textbf{Artificial + 100/200/500 real samples (Disfl-QA):} We combine the samples of real examples of Disfl-QA with the corresponding generated artificial disfluencies.
\end{itemize}

All the models were evaluated on the original test sets of Switchboard and Disfl-QA. Results in Table~\ref{tab:low_resource} demonstrate that even when no real disfluent data are used, we can achieve an adequate performance of more than 84\% in terms of F1 for both datasets, only with artificial disfluencies. This finding is promising, as it suggests that the proposed method can be successfully used for low-resource languages, where annotated datasets for disfluency detection do not typically exist. In addition, we can achieve better performance on the Switchboard dataset when only artificial disfluencies are used compared to the model fine-tuned with 1000 samples of real data, which achieves 82.41\% in terms of F1. The same conclusions can be drawn for the Disfl-QA dataset, where better performance is achieved with only artificial disfluencies compared to the model fine-tuned with 100 disfluent samples with only 78.09\% F1. 

\begin{table*}[!t]
\caption{Results on sequence-tagging models fine-tuned on limited samples of both Switchboard and Disfl-QA dataset, with or without artificial generated disfluencies. Average token-based precision (Prec), recall (Rec) and f-measure (F1) (\%) is reported.}
\tiny
\label{tab:low_resource}
\resizebox{0.99\textwidth}{!}{%
\begin{tabular}{cccc|cccc} \hline \\
\multicolumn{4}{c}{\textbf{Switchboard}} &  \multicolumn{4}{c}{\textbf{Disfl-QA}} \\
& \textbf{Prec} & \textbf{Rec} & \textbf{F1} & & \textbf{Prec} & \textbf{Rec} & \textbf{F1}  \\ \hline
1K real samples & 92.60 & 76.46 & 82.41 & 100 real samples    & 86.64 & 75.30 & 78.09 \\
2K real samples & 91.56 & 84.42 & 87.61 & 200 real samples   & 88.10 & 82.40 & 84.48 \\ 
5K real samples  & 91.23 & 84.63 & 87.60 & 500 real samples   & 91.31 & 89.65 & 90.41 \\ \hline
only artificial samples & 95.93 & 78.39 & 84.84 & only artificial samples & 87.20 & 82.47 & 84.23 \\ 
artificial + 1K real samples & 96.98 & 80.40 & 86.76 & artificial + 100 real samples & 88.83 & 85.93 & 87.16 \\ 
artificial + 2K real samples & 96.31 & 82.23 & 87.85 & artificial + 200 real samples & 90.03 & 87.72 & 88.75 \\
artificial + 5K real samples & \textbf{98.08} & \textbf{84.81} & \textbf{89.71}  & artificial + 500 real samples & \textbf{92.34} & \textbf{91.01} & \textbf{91.63}  \\  
\hline
\end{tabular}}
\end{table*}

Additional gains are observed when artificial data are combined with samples of real data. We can notice that all the models fine-tuned with combined data outperform all the models that are fine-tuned only with the corresponding samples of real data in terms of F1 for both datasets. Furthermore, the best performance for the Switchboard dataset is achieved, when 5,000 samples are combined with artificial disfluent data, with an increase of more than 6 points in terms of precision (98.08\% compared to 91.23\%) and more than two points in terms of F1 (89.71\% compared to 87.60\%). Similarly, we obtain the best performance for the Disfl-QA dataset, when 500 samples are inserted into artificial disfluent data, with an increase of more than 5 points in terms of precision (92.34\% compared to 86.64\%) and more than 3 points in terms of F1 (91.63\% compared to 78.09\%). In all cases, a performance increase is noticed, which further demonstrates the capability of the proposed method to achieve strong performance with no or only a few annotated data.

\subsubsection{Artificial data with real data}
In many cases, the insertion of artificial disfluencies directly into real data for training can lead to a performance decrease due to distribution shifts between the real and artificial data~\citep{dong2019adapting, yang2020planning}. However, in this experiment, we demonstrate that the proposed method does not alter the distribution of real data, by generating realistic disfluencies, which increases the accuracy of existing models. In particular, we investigate the effect of artificial disfluencies when inserted directly into the full Switchboard and Disfl-QA training sets. We extract all the fluent sentences of both datasets and we generate all the different types of artificial disfluencies, namely repetitions, replacements and restarts. We insert 3,000 and 10,000 artificial repetitions, replacements and restarts for Disfl-QA and Switchboard, respectively. Then, we fine-tune both sequence tagging and translation-based models on the original training datasets with different types of inserted disfluencies. Results of inserted repetitions, replacements, and restarts on the Disfl-QA and Switchboard dataset are shown in Table~\ref{tab:inserted_disfluencies_sw} and~\ref{tab:inserted_disfluencies_disfl_qa}, respectively. All the models were evaluated on the original test set of Disfl-QA and Switchboard, respectively.

\begin{table}[!hbt]
\caption{Results on sequence-tagging and traslation-based models fine-tuned on Disfl-QA with inserted artificial repetitions (rep.), replacements (repl.) and restarts (res.)}
\label{tab:inserted_disfluencies_disfl_qa}
\resizebox{0.99\columnwidth}{!}{%
\begin{tabular}{cccc|c} 
\hline \\
& \multicolumn{3}{c}{\textbf{Sequence Tagging}} & \textbf{Translation-based}\\
     & \textbf{Prec}  & \textbf{Rec}   &  \textbf{F1}  & \textbf{BLEU}  \\ \hline
original                        & 96.07 & 96.02 & 96.04 &  94.95 \\
original + rep.                 & 96.38 & 96.24 & 96.31 &  94.90 \\ 
original + repl.                & 96.25 & 96.10 & 96.18 &  94.96 \\
original + rest.                & \textbf{96.54} & \textbf{96.41} & \textbf{96.69} &  94.91 \\ \hline 

original + all                  & 96.32 & 96.13 & 96.22 &  \textbf{95.10}  \\
\hline
\end{tabular}}
\end{table}

\begin{table}[!hbt]
\caption{Results on sequence-tagging and translation-based models fine-tuned on Switchboard with inserted artificial repetitions (rep.), replacements (repl.) and restarts (res.)}
\label{tab:inserted_disfluencies_sw}
\resizebox{0.99\columnwidth}{!}{%
\begin{tabular}{cccc|c} 
\hline \\
& \multicolumn{3}{c}{\textbf{Sequence Tagging}} & \textbf{Translation-based}\\
     & \textbf{Prec}  & \textbf{Rec}   &  \textbf{F1}  & \textbf{BLEU}  \\ \hline
original      & 96.45 & 91.63 & 93.89 &  91.18 \\
original + rep. & 96.41 & \textbf{92.77}  & \textbf{94.50} & \textbf{91.24} \\ 

original + repl.       & 96.11 & 92.31  & 94.12 & 91.17 \\
original + rest. & 92.32 & 92.28 & 92.30 & 90.50  \\ 
 \hline
original + rep. + repl.        & \textbf{96.57}     & 92.05     & 94.18 &  91.21  \\
\hline
\end{tabular}}
\end{table}

Results in Table~\ref{tab:inserted_disfluencies_disfl_qa} indicate that when synthetic data of any type are inserted, the performance of all the sequence tagging models is increased, contrary to when no synthetic data is used. Similar results are observed for the majority of the translation-based models, where inserted replacements and restarts lead to an increased performance as well. The highest accuracy is achieved when artificial restarts are inserted for the sequence tagging model with 96.69\% compared to 96.04\% in terms of F1 and when both artificial replacements and restarts are inserted for the translation-based model with 95.10\% compared to 94.95\% in terms of BLEU score. This finding suggests that the generated artificial disfluencies can be effectively combined with existing datasets that contain real disfluencies without altering their distributions.

Similar results are obtained on the Switchboard dataset as shown in Table~\ref{tab:inserted_disfluencies_sw}, where the majority of fine-tuned models with inserted disfluencies achieve an increased accuracy. However, contrary to Disfl-QA, we noticed a slight drop in performance when artificial restarts are inserted into the original dataset. This can be attributed to the nature of Switchboard examples, which are typically incomplete sentences. In some cases, this can be tricky for the restart algorithm, leading to misclassified fluent sentences. Therefore, even though the drop in performance is not significant, artificial restarts should be generated with caution. Additionally, the best performance is obtained when both repetitions and replacements are inserted with 94.18\% in terms of F1 and 92.05\% in terms of recall, compared to the 93.89\% and 91.63\% of the model fine-tuned on the original dataset. For the translation-based model, we achieve the best accuracy when only artificial repetitions are inserted with 91.24\% compared to 91.18\% in terms of BLEU score. Since translation-based is a more challenging task, we expect a slighter increase in contrary to the sequence tagging task. In addition, the combination of both repetitions and replacement still increases the accuracy of the original model with 91.21\% compared to 91.18\% BLEU score. 

\section{Conclusions and Future Work}
\label{sec:conclusion}
We proposed LARD, a method for automatically generating artificial disfluencies. The proposed method consists of three different algorithms for handling the three most common types of disfluencies: repetitions, replacements, and restarts. LARD can be directly used without any labeled data to successfully train disfluency detection models. Our empirical evaluation showed the effectiveness of the proposed method in a low resource setup, when none or few annotated data are available. In addition, our analysis confirmed that LARD can generate realistic disfluencies which do not alter the distribution of data that contain real disfluencies.  

Future work could examine the effect of the proposed method for generating more than one disfluency per sentence. Furthermore, it would be very interesting to extend the proposed method to other languages, where disfluent data are not readily available. 

\section*{Acknowledgments}
This work is partially funded by the European Commission as part of its H2020 Programme, under the contract number 870930-IA (WELCOME).

% {\appendix[Proof of the Zonklar Equations]
% Use $\backslash${\tt{appendix}} if you have a single appendix:
% Do not use $\backslash${\tt{section}} anymore after $\backslash${\tt{appendix}}, only $\backslash${\tt{section*}}.
% If you have multiple appendixes use $\backslash${\tt{appendices}} then use $\backslash${\tt{section}} to start each appendix.
% You must declare a $\backslash${\tt{section}} before using any $\backslash${\tt{subsection}} or using $\backslash${\tt{label}} ($\backslash${\tt{appendices}} by itself
%  starts a section numbered zero.)}


% \section{References Section}
% You can use a bibliography generated by~\citep{sezgin2020readiness} BibTeX as a .bbl file.
%  BibTeX documentation can be easily obtained at:
%  http://mirror.ctan.org/biblio/bibtex/contrib/doc/
%  The IEEEtran BibTeX style support page is:
%  http://www.michaelshell.org/tex/ieeetran/bibtex/
 
\bibliography{lrec2022-example}
\bibliographystyle{IEEEtranN}

% \newpage

\section{Biography Section}
\vspace{-33pt}

 
% \vspace{-11pt}

% \bf{If you include a photo:}
\begin{IEEEbiography}[{\includegraphics[width=1in,height=1.25in,clip,keepaspectratio]{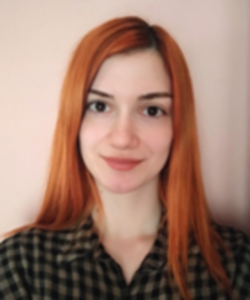}}]{Tatiana Passali} received the B.S. degree in Applied Informatics from the University of Macedonia, Thessaloniki, Greece, in 2019, and the M.S. degree in 2021 in data and web science from Aristotle University of Thessaloniki (AUTH), Thessaloniki, Greece, where she is currently pursuing the Ph.D. degree in natural language processing. She is currently a research associate with Center of Research and Tenchology Hellas (CERTH) in Greece. Her current research interests include deep learning, natural language processing, automatic summarization and dialogue systems.
\end{IEEEbiography}

\begin{IEEEbiography}[{\includegraphics[width=1in,height=1.25in,clip,keepaspectratio]{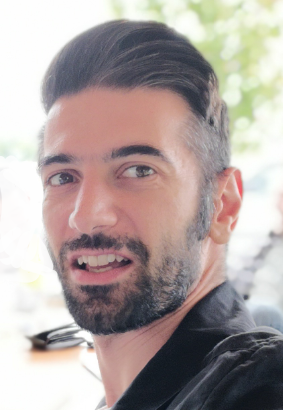}}]{Thanassis Mavropoulos}
received his PhD degree in Computational Linguistics from the Aristotle University of Thessaloniki, Hellas and is working as a postdoctoral research fellow at the Information Technologies Institute (ITI) of the Centre for Research and Technology Hellas (CERTH) since 2016. His research interests include Natural Language Processing and Semantics. He has successfully participated in several European projects related, among others, to domains that relate to Health, Cybersecurity, Migration, and Virtual Agents. His scientific work has been featured in refereed journals and international conferences; he has also served as a reviewer in various journals and conferences, while also being involved in some as a program committee member.
\end{IEEEbiography}

\begin{IEEEbiography}[{\includegraphics[width=1in,height=1.25in,clip,keepaspectratio]{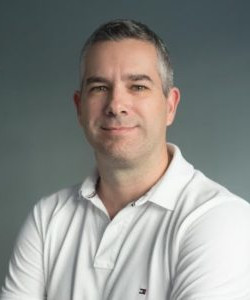}}]{Grigorios Tsoumakas} received a degree in Informatics from the Aristotle University of Thessaloniki (AUTH) in 1999, an MSc in Artificial Intelligence from the University of Edinburgh in 2000 and a PhD in Informatics from AUTH in 2005. He is an Associate Professor of Machine Learning and Knowledge Discovery at the School of Informatics of AUTH since 2020, where he has also served as Assistant Professor (2013 – 2020) and Lecturer (2007 – 2013). His research expertise focuses on supervised machine learning (ensemble methods, multi-target prediction) and document intelligence (semantic indexing, keyphrase extraction, summarization). He has published more than 140 articles and according to Google Scholar his work has received more than 16K citations and his h-index is 47. His honors include receiving the European Conference on Machine Learning and Principles and Practice of Knowledge Discovery in Databases (ECML PKDD) 10-Year Test of Time Award in 2017 and the Marco Ramoni best paper award at the 19th International Conference on Artificial Intelligence in Medicine (AIME) 2021. Dr. Tsoumakas is a senior member of IEEE and ACM. He is an advocate of applied research that matters and has worked as a machine learning engineer, researcher and consultant in several national, international, and private sector funded R\&D projects. He is academic co-founder of Medoid AI, a spin-off company of AUTH established in 2019.
\end{IEEEbiography}

\begin{IEEEbiography}[{\includegraphics[width=1in,height=1.25in,clip,keepaspectratio]{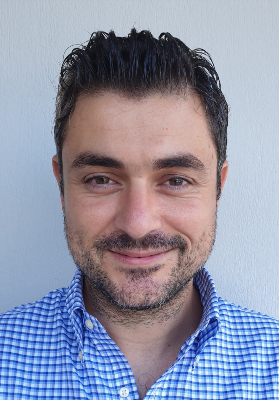}}]{Georgios Meditskos} received in 2009 the PhD degree in Informatics from Aristotle University of Thessaloniki in Greece for his dissertation on "Semantic Web Service Discovery and Ontology Reasoning using Entailment Rules". He also holds an MSc and a BSc degree from the same department. In 2012 he joined the Information Technologies Institute (ITI) of the Center for Research and Technology Hellas (CERTH) as a postdoctoral researcher. Since July 2021 he is an Assistant Professor in the School of Informatics at the Aristotle University of Thessaloniki, Greece. He is the author of more than 100 publications in refereed journals and international conferences. He has actively participated in more than 10 European and National projects relevant to Cultural Heritage, Health, Cybersecurity, Disaster Management, Virtual Environments and Agents. His research interests revolve around Symbolic AI, and more specifically on knowledge representation and reasoning in the Semantic Web (RDF/OWL, rule-based ontology reasoning, combination of rules and ontologies), Knowledge Graphs and context-based multi-sensor reasoning and fusion in Pervasive Environments.
\end{IEEEbiography}

\begin{IEEEbiography}[{\includegraphics[width=1in,height=1.25in,clip,keepaspectratio]{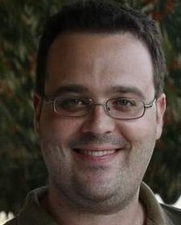}}]{Stefanos Vrochidis} received the Diploma degree in Electrical Engineering from Aristotle University of Thessaloniki, Greece, the MSc degree in Radio Frequency Communication Systems from University of Southampton and the PhD degree in Electronic Engineering from Queen Mary University of London.  Currently, he is a Senior Researcher (Grade C) with the Information Technologies Institute of the Centre for Research and Technology Hellas (ITI-CERTH) and the Head of the Multimodal Data Fusion and Analytics (M4D) Group of the Multimedia Knowledge and Social Media Analytics Lab. His research interests include multimedia analysis and retrieval, multimodal fusion, computer vision, multimodal analytics, artificial intelligence, as well as media \& arts, environmental and security applications. Dr. Vrochidis has participated in more than 60 European and National projects and has been member of the organization team of several conferences and workshops. He has edited 3 books and authored more than 220 related scientific journal, conference and book chapter publications. He has served as a reviewer in several international Journals and as Technical program committee in well reputed conferences and workshops.
\end{IEEEbiography}

\vfill



% \vfill

\end{document}